\documentclass[conference,compsoc]{IEEEtran}
\ifCLASSINFOpdf
\else
\fi
\usepackage{ulem}
\usepackage{amsmath}
\usepackage{mathrsfs}
\usepackage{hyperref}       
\usepackage{url}            

\usepackage{graphicx}
\usepackage{graphics}
\graphicspath{{images/}}
\usepackage{booktabs}
\itshape

\small
\newcommand\embf{\textbf}

\usepackage[numbers,sort&compress]{natbib}

\usepackage{flushend}

\begin{document}


%
\title{High-Quality Face Image Super-Resolution Using \\ Conditional Generative Adversarial Networks}

\author{\IEEEauthorblockN{Huang Bin and Chen Weihai and Wu Xingming}
\IEEEauthorblockA{School of Automation Science \\and Electrical Engineering,
Beihang University\\
XueYuan Road No.37, HaiDian District, Beijing, China\\
Email: marshuangbin@buaa.edu.cn\\
Email:whchenbuaa@126.com\\
Email:wxmbuaa@163.com}
\and
\IEEEauthorblockN{Lin Chun-Liang}
\IEEEauthorblockA{Department of Electrical Engineering,
\\ National Chung Hsing University\\
145 Xingda Rd., South Dist.\\Taichung City 402, Taiwan \\
Email:chunlin@dragon.nchu.edu.tw}

}


%


\maketitle

\begin{abstract}
We propose a novel single face image super-resolution method, which named \textit{Face Conditional Generative Adversarial Network}(FCGAN), based on boundary equilibrium generative adversarial networks. Without taking any facial prior information, our method can generate a high-resolution face image from a low-resolution one. Compared with existing studies, both our training and testing phases are end-to-end pipeline with little pre/post-processing. To enhance the convergence speed and strengthen feature propagation, skip-layer connection is further employed in the generative and discriminative networks. Extensive experiments demonstrate that our model achieves competitive performance compared with state-of-the-art models.
\end{abstract}


%
\IEEEpeerreviewmaketitle

\section{Introduction}
Single image super resolution(SISR), a greatly challenging task of computer vision and machine learning, attempts to reconstruct a high-resolution(HR) image from a low-resolution(LR) image. Super resolution(SR) is commonly divided into two categories based on their tasks, namely generic image SR and class-specific image SR. The former takes little class information into account, which aims to recover any kinds of high resolution image from corresponding low-resolution image. In general, the latter usually refers to face image super resolution or face hallucination if the class is face.

Face image super resolution or face hallucination\citep{Jia2011Fast, zhu2016deep, wang2014comprehensive, li-PR2014face, autee2015review, Jiang2016Noise-TCYB, jin2015robust, su2016supervised, W2016DeepJFHR} is an important branch of super-resolution(SR). The great distinction between the both techniques is that face hallucination always employs typical facial priors (eg. face spatial configuration and facial landmark detection) with strong cohesion to face domain concept. More realistic and sharper details, which plays a crucial role in intelligence surveillance\cite{Jia2011Fast, wang2014comprehensive} and face recognition\cite{W2016DeepJFHR}, are taken by HR face images than corresponding LR images. Due to long distance imaging, the limitations on storage and low-cost electronic imaging systems, LR images appear in many cases instead of HR images. Thus, SR has turned out to be an active research filed in the past few years.

Face image SR is an ill-posed problem (as same as generic image SR), for which it needs to recover 16 pixels (for $4\times$ upscaling factors) from each given pixel. While, recent years have witnessed a tremendous growth of research and development in the field, in particular using learning-based methods.
\begin{figure}[h]
  \centering
  \includegraphics[width = 8.5cm]{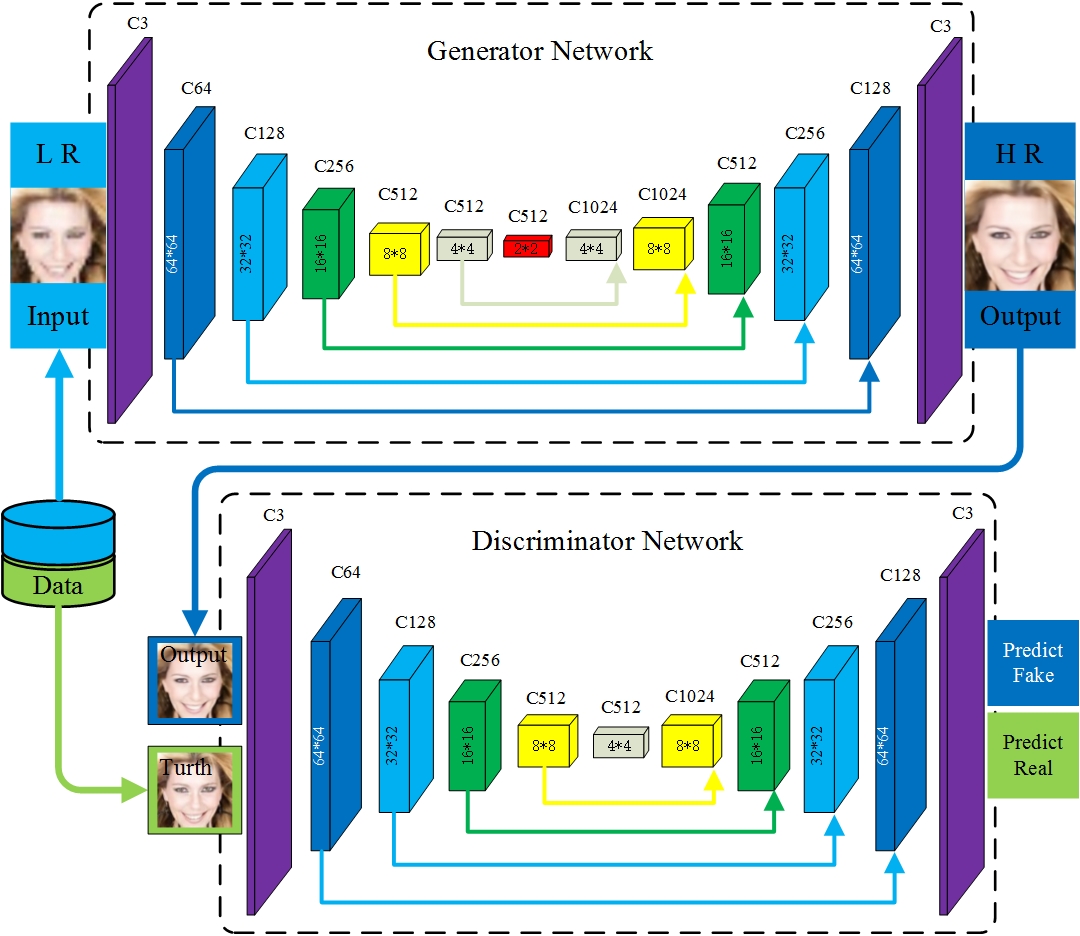}
  \caption{The pipeline of FCGAN. The architecture of generator and discriminator network with corresponding filter size and output channels(C) for each convolutional layer. In the testing phase, only the generator network is employed and the discriminator network does not work. }
  \label{Fig:architecture}
\end{figure}

In this paper, we propose a HR face image framework ($4\times$ upscaling factors) based on boundary equilibrium generative adversarial network(BEGAN)\cite{David-BEGAN2017}. In order to adapt BEGAN for SR task, single low-resolution face image is considered as the prior condition to generate a high-resolution one. So, we refer to the framework as \textit{Face Conditional Generative Adversarial Network}(named FCGAN for short hereafter). Our proposed method does not utilize any priors on face structure or face spatial configuration. In addition, it is also an end-to-end solution to generate HR face images without need any pre-trained model. We perform extensive experiments, which demonstrates that our method not only achieves high Peak Signal to Noise Ratio(PSNR), but also improves actual visual quality.

Overall, the contributions of this paper are mainly in three aspects:

\begin{itemize}
  \item We propose a novel end-to-end method (FCGAN), with $4\times$ upscaling factors, to learn mapping between low-resolution single face images to high-resolution one. The method can robustly generate a high-quality face image from low-resolution one.
  \item To the best of our knowledge, our method is the first attempt to develop BEGAN\cite{David-BEGAN2017} to generate HR face images from low-resolution ones regardless of pose, facial expressions variation, face alignment and lighting. Our model considers a low-resolution image $I^{LR}$ as the input instead of random noise.
  \item We introduce the pixel-wise $L_1$ loss function to optimize the generative and discriminative models. Compared with state-of-the-art models, extensive experiments show that FCGAN achieve competitive performance on both visual quality and quantitative analysis.

\end{itemize}


\section{Related work}

In general, image SR methods can be classified into three categories: interpolation methods, reconstruction-based methods, and example (learning)-based methods. Among them, due to the simply pipeline and excellent performance, the example-based methods\cite{zhu2016deep, Jiang2016Noise-TCYB, jin2015robust, W2016DeepJFHR, Christian-SRGAN-CVPR2017, dong2016accelerating, Dong-He-PAMI2016Image, Jiang2016SRLSP, Kim_2016_DRCN, LapSRN_CVPR2017, Yu-URGAN-ECCV2016} achieve explosive development in the past years. In this section, we will also mostly focus on discussion example-based methods.

\subsection{generic image SR}
In the past few years, Deep convolutional neural networks(DCNNs) have demonstrated outstanding performance in single image SR. Dong et al.'s work\cite{Dong-He-PAMI2016Image} first extend CNN to the field of image SR and demonstrate that deep learning can achieve higher quality image than other learning-based methods. The authors design a simple fully convolutional neural network that directly learns an end-to-end mapping between low-resolution and high-resolution images. Furthermore, they point out that the three convolutional layers can be abstracted into patch extraction and representation, non-linear mapping and reconstruction, respectively. Several excellent models\cite{dong2016accelerating, Kim_2016_DRCN, LapSRN_CVPR2017} are presented to improve the performance based on CNNs.

In general, the more layers the CNN model has, the better the model performance, but the deep model convergence speed becomes a critical issue during training. However, in Kim's work\cite{Kim_2016_DRCN}, named VDSR for short, the very deep convolutional network was proposed based on residual-learning\cite{He2015Deep}, which can effectively strengthen the transfer of the gradient and enhance the convergence speed. In their model, the magnitude of convolutional layers is up to 20, while the model presented in \cite{Dong-He-PAMI2016Image} only has 3 layers. Compared with Dong's work\cite{Dong-He-PAMI2016Image}, however, VDSR achieves better performance not only on image quality, but also on the running time. Recently, Lai et al.\cite{LapSRN_CVPR2017} proposed a Laplacian Pyramid Super-Resolution Network(LapSRN) based on a cascade of convolutional neural networks(CNN). The network progressively predicts the sub-band residual in a coarse-to-fine fashion and is trained with a robust Charbonnier loss function to reconstruct the high-frequency information.

Different from the previous works, generative adversarial network(GAN) is one of the most common methods\cite{David-BEGAN2017, Christian-SRGAN-CVPR2017, Yu-URGAN-ECCV2016, Goodfellow2014GAN} to adapt for SR. Due to the discriminative network, GAN-based methods can generate HR images with much sharper details than other generative models\cite{kingma2013auto, denton2015deep}. In order to reconstruct more realistic texture details with large upscaling factors, Christian et al.\cite{Christian-SRGAN-CVPR2017} proposed a deep residual network with the perceptual loss function which consists of an adversarial loss and a content loss. Specifically, the authors calculated the content loss based on high-level feature maps of VGG network\cite{Simonyan2015Very} instead of MSE(the mean squared error).

\subsection{Face image SR}

Face image SR, also called face hallucination, is an important branch of SR. Due to face inherently possesses specific spacial configuration (e.g., facial landmarks localization). So, it is very obvious that facial features and landmarks can be extracted as guidance of prior to recover HR face images. For example, Jiang et al.\cite{Jiang2016Noise-TCYB, Jiang2016SRLSP} proposed a face image SR method using smooth regression with local structure prior(SRLSP). The authors consider the relationship between the LR image patch and the hidden HR pixel information as local structure prior, which is then used to recover HR face image from the LR one. Because of the overlap patch mapping, the above method is time consuming.

However, Zhu et al.\cite{zhu2016deep} pointed out that is a chicken-and-egg problem - HR face image is better recovered by face spatial configuration, while the latter requires a higher resolution face image.  To address the problem, the authors proposed the \textit{Cascaded Bi-Network}(CBN) with alternatingly optimizing two branch networks(face hallucination and dense correspondence filed estimation). The latter branch is capable of reconstructing and synthesizing latent texture details from the LR face image.

The methods based on GAN architecture can also applied to generate HR face image from LR one. Different from aforementioned methods\cite{zhu2016deep, Jiang2016Noise-TCYB, Jiang2016SRLSP}, Yu et al. \cite{Yu-URGAN-ECCV2016} presents a discriminative generative network, without capturing any prior information, to recover HR face images with high upscaling factors($8\times$). However, there are two drawbacks with this method. One is that the face train set require frontal and approximately aligned, the other is that the generative face images are sensitive to rotations.


\section{Proposed method}
\label{gen_inst}

The aim of Single Image Super Resolution(SISR) is to estimate the mapping from lower-resolution input image $I^{LR}$ to high-resolution output images $I^{HR}$. Here the $I^{LR}$ downsample from corresponding $I^{HR}$ in a general way. Philip et at.'s\cite{pix2pix} research shows that conditional generative adversarial networks\cite{mirza2014conditionalGAN} are a promising approach for a variety of image-to-image translation tasks. Inspired by their works\cite{mirza2014conditionalGAN, pix2pix}, we considered $I^{LR}$ to $I^{HR}$ as a conditional transition task, namely $I^{LR}$ is the condition to generate $I^{HR}$. Furthermore, our proposed FCGAN method extends the Wasserstein distance\cite{Arjovsky2017TowardsWGAN, Arjovsky2017WGAN, David-BEGAN2017} to optimize the networks in our model.

\subsection{Model architectures}
The structure of our model is shown in figure \ref{Fig:architecture}. We adapt our generator and discriminator architecture from the U-Net\cite{ronneberger2015U-net} which is an encoder-decoder with skip connections between mirrored layers in the encoder and decoder stacks. The skip layer connections have been used in many solutions\cite{ronneberger2015U-net, huang2016Densenet, orhan2017skip, Densenetsemantic} in the filed of Deep Convolutional Neural Network(DCNN).

We design the network architecture around the following considerations. The skip connections can strengthen feature propagation and encourage feature reuse between the two connected layers. If not use skip connections, the information (taken by the previous feature map) will missing progressively when passed through a series of layers, and the convergence speed of the model will be also slow down sharply in the training phase.

The architecture of generator G: $R^{N_x} \rightarrow R^{N_y}$ is a fully convolutional neural network to generate HR image corresponding with the input LR image. $N_x = H\times W \times C$ is short for the dimensions of $x$ where $H,W,C$(for RGB image $C=3$) are height, width and colors, respectively. In order to make sure the dimensions of connection features in different layers to be the same, we implement the convolution with the kernel size of $4\times4$ in each layer and set $stride = 2$  to reduce the feature maps' dimensions. LeakyReLU activation($\alpha = 2$) is used, and pooling operation avoid to use throughout the network. The generator network G illustrated in the upper section of figure \ref{Fig:architecture} contains six downsampling convolutional layers and six upsampling convolutional layers with a decreasing/increasing factors of 2. In short, the structure of G can be simply referred to as the following pipeline: $128\times128\times3(input)\rightarrow 64\times64\times64 \rightarrow 32\times32\times128 \rightarrow 16\times16\times256 \rightarrow 8\times8\times512 \rightarrow 4\times4\times512 \rightarrow 2\times2\times512 \rightarrow 4\times4\times1024 \rightarrow 8\times8\times1024 \rightarrow 16\times16\times512 \rightarrow 32\times32\times256 \rightarrow 64\times64\times128 \rightarrow 128\times128\times3(output)$.

The architecture of discriminator D: $R^{N_{ry}} \rightarrow R^{N_{ry}}$, where $R^{N_{ry}}$, having the dimensions of ($H \times W \times 2C$), is grouped by the output(generative SR image simple) of G and corresponding real SR image sample. As showing in the bottom section of figure \ref{Fig:architecture}, the architecture of D is similar with G. There are only two crucial distinguishable points between G and D network, one is the input/output dimensions, the other is that D has only ten convolutional layers(five downsampling and upsampling layers).

\subsection{Loss function}

Typical GANs try to capture training data distribution\cite{Goodfellow2014GAN}: generator G learns the distribution $p_g$ over data $x$ to generate fake data $G(x)$, and discriminator D distinguishes the distribution of a sample whether belongs to real or fake data. Inspired by \cite{Arjovsky2017WGAN, zhao2016EBGAN}, our method attempts to match the loss distribution directly at the pixel level. Thus, in our model, we use the $L_1$ norm to measure the loss error between the generative sample $G(z)$ and the corresponding sample $x$. Motivate by David et al.\cite{David-BEGAN2017}, we adapt original GAN\cite{Goodfellow2014GAN} loss function as pixel-wise $L_1$ norm to optimize the generator and discriminator network loss function. The generator $L_1$ norm loss function as shown following equation \ref{eq:pixel_loss}.
\begin{align}\label{eq:pixel_loss}
    \mathcal{L}(I) = |I^{HR}-G(I^{LR})|
\end{align}

As the research of BEGAN\cite{David-BEGAN2017} shown, the image-wise loss distribution is approximately normal under condition of a sufficient substantial number of pixels. Thus, the objective function can further simplify to the equation \ref{eq:GD_loss}, where $x$ is real HR face sample, $z$(input of G) is the LR face sample, $y$ is the fake HR face image (output of G) generated by G with $z$, and $\mathcal{L}_D$ represents the global loss of D. In addition, in the equation \ref{eq:preD_loss}, where $\mathcal{L}_{D_r}$ represents the discriminator loss with real sample, $\mathcal{L}_{D_f}$ represents the discriminator loss with fake sample generated by G. Given the discriminator and generator parameters $\theta_D$ and $\theta_G$, which updated by minimizing the losses $\mathcal{L}_D$ and $\mathcal{L}_G$.
\begin{align}\label{eq:preD_loss}
y & = G(x;\theta_D) \\
\mathcal{L}_{D_r} & = \mathcal{L}(D(x;\theta_D)-x) \\
\mathcal{L}_{D_f} & = \mathcal{L}(D(y;\theta_D)-y) \notag \\
& = \mathcal{L}(D((G(z;\theta_G))-G(z;\theta_G));\theta_D)\\
\mathcal{L}_D & = \mathcal{L}_{D_r} - \mathcal{L}_{D_f}
\end{align}
\begin{equation}\label{eq:GD_loss}
  \begin{cases}
   \mathcal{L}_D = \mathcal{L}_{D_r} - \mathcal{L}_{D_f}, & \text{for } \theta_D \\
   \mathcal{L}_G = \mathcal{L}(G(z)-x), & \text{for } \theta_G
  \end{cases}
\end{equation}

To maintain the optimization level between the generator G and discriminator D, we finally use the equilibrium algorithm\cite{David-BEGAN2017} as shown in the equation \ref{eq:Equilibrium}. If not, the parameters of generative network may be optimized in a high level, but the discriminator is still in poor level. The essential idea of the algorithm is a form of closed-loop feedback control to maintain the balance of the whole training process. We set $\gamma = 0.5$, $\lambda = 0.001$ in our experiments.
\begin{align}\label{eq:Equilibrium}
\begin{cases}
  \mathcal{L}_D = \mathcal{L}_{D_r} - k_t \mathcal{L}_{D_f} \\
  k_{t+1} = k_t + \lambda_k(\gamma \mathcal{L}_{D_r} - \mathcal{L}_G)
\end{cases}
\end{align}

Furthermore, we employ $\mathcal{M}_c$ \cite{David-BEGAN2017} (as shown in the equation \ref{eq:measure}) to measure the convergence level of our model.
\begin{align}\label{eq:measure}
\mathcal{M}_c = \mathcal{L}_{D_r} + |\gamma \mathcal{L}_{D_r}-\mathcal{L}_G|
\end{align}

These equations, while similar to those from BEGAN, have two important differences:
\begin{itemize}
  \item The input of generator, which not a random vector sample, is LR face image. We regard the input as a condition for generating HR face image. Thus, our method can control the generative face.
  \item We use $L_1$ norm as the pixel-wise loss functions of generator, as the equation \ref{eq:GD_loss} shown.
\end{itemize}

\section{Experiments}

We trained our model using Adam with the learning rate of 0.0001. After 10 iterations training with CelebA\cite{liu2015Celeba} face dataset, our model converged to its final state, which spend about 120 minutes in the machine (one NVIDIA TITAN X GPU, 12G). In order to demonstrate the performance of FCGAN, we will compare our results to the state-of-the-art methods\cite{dong2016accelerating, LapSRN_CVPR2017, pix2pix} and evaluate it qualitatively and quantitatively in the section 4.2.

\begin{figure}[h]
  \centering
  \includegraphics[width = 8.5cm]{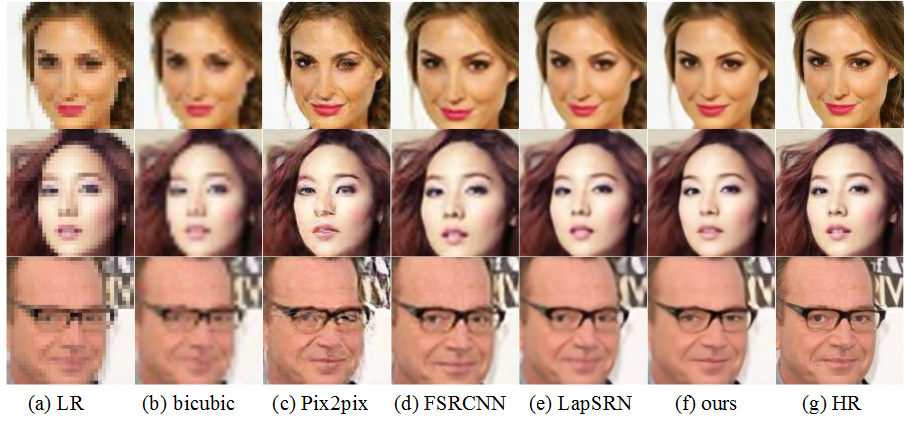}
  \caption{Comparison with the state-of-the-art methods training with CelebA dataset. (a) LR images. (b) Bicubic interpolation. (c) Philip et al.'s method\cite{pix2pix}. (d) Dong et al.'s method\cite{dong2016accelerating}. (e) Lai et al.'s method\cite{LapSRN_CVPR2017}. (f) FCGAN(ours). (g) Original HR images.}
  \label{Fig:comparison}
\end{figure}

\subsection{Setup}
\embf{Datasets.} CelebA\cite{liu2015Celeba} is a large-scale face attributes dataset with more than 200k celebrity images, each with 40 attribute annotations. The dataset covers large pose variations and background clutter. Before training our proposed model with CelebA dataset, we cropped the images and resize them to $128\times128$. We randomized the cropped images, and then used more than 180k images for training, 10k images for validation, 10k images for testing.

\embf{Set up LR datasets.} Firstly, we downsample the HR images ($128 \times 128$) to the resolution of $32\times32$ pixels (LR images). Then, we employ bicubic interpolation algorithm to generate interpolative images (named BHR, with the size of $128\times128$), and finally constructed the BHR and HR images to the input-output pairs($b_{i},h_{i}$). So, the input and output images of FCGAN are same size of $128\times128$ with three color channels.

\subsection{Experimental results and analysis}

\begin{figure}[h]
  \centering
  \includegraphics[width = 8.5cm]{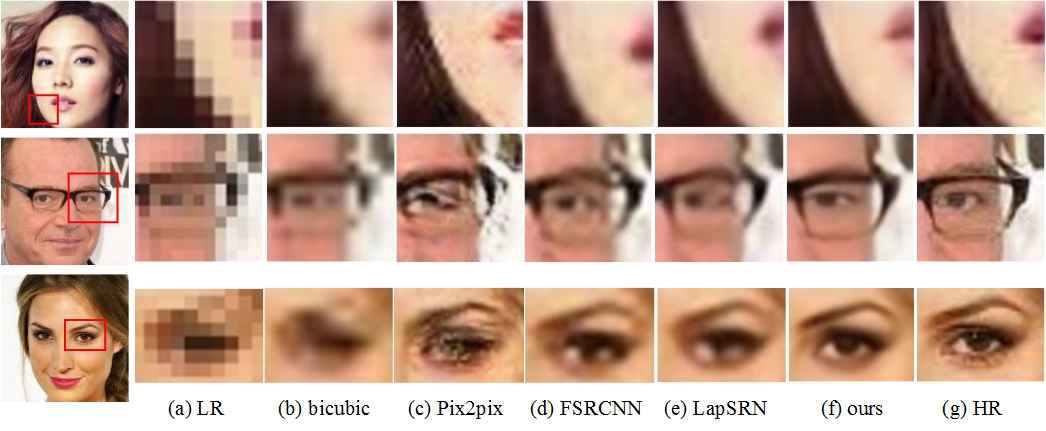}
  \caption{Visual comparisons on local details}
  \label{Fig:local_details}
\end{figure}

In this section, we compare our FCGAN with currently state-of-the-art SR methods. In order to make a fair comparison, we retrain all other algorithms with the dataset CelebA. We report the qualitative results in figure \ref{Fig:comparison}, and provide the quantitative results in table \ref{tab:PSNR}. Furthermore, the figure \ref{Fig:local_details} shows the more clearly local details of the generative HR images. As can be seen from the results, our FCGAN method has significant advantages over other methods.

\begin{table}
  \caption{Quantitative comparisons on the CelebA dataset}
    \label{tab:PSNR}
  \begin{tabular}{ccccccc}
    \toprule
    & LR & bicubic & pix2pix & FSRCNN & LapSRN & ours \\
    \midrule
  PSNR & 29.46 & 31.25 & 30.27 & 31.92 & 32.13 & 32.42 \\
    \bottomrule
\end{tabular}
\end{table}

\begin{figure}[h]
  \centering
  \includegraphics[width = 9cm]{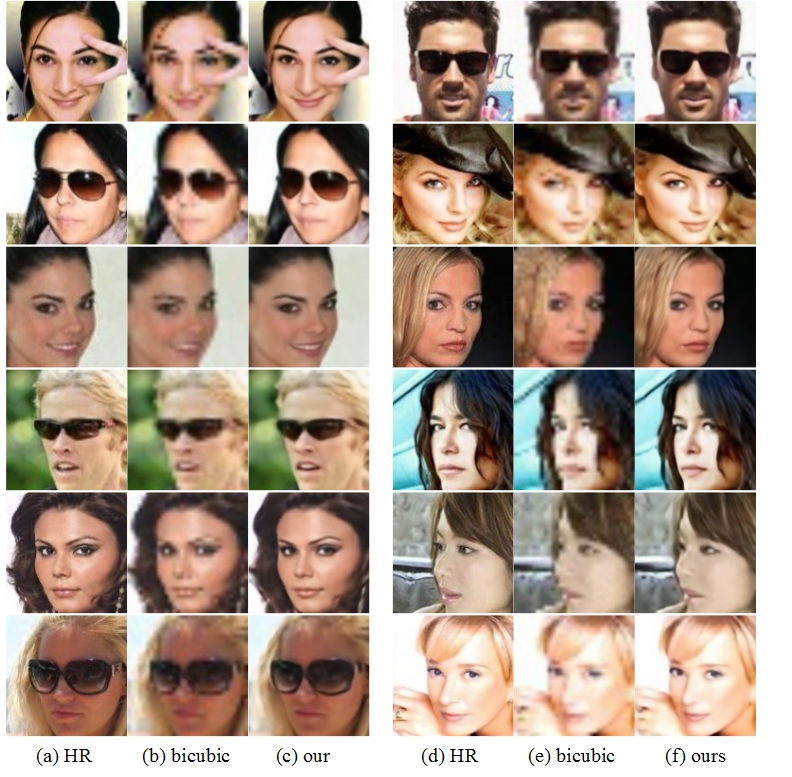}
  \caption{Qualitative HR face images generated by our method with $4\times$ upscaling factors}
  \label{Fig:gene_results}
\end{figure}

As shown in figure \ref{Fig:gene_results}, more results generated by our FCGAN method are listed. It is worth pointing out that FCGAN can robustly generate high-quality face images ($4 \times$) regardless of facial expression, pose, illumination, occlusion(wearing glasses or hat), and other factors.

\section{Conclusion and future work}
In this paper, we have proposed a novel SR method ($4\times$ upscaling factors) to generate a HR face image from LR one, namely \textit{Face Conditional Generative Adversarial Network} (FCGAN). In this model, the LR image, instead of random noise, is considered as a controller to generate a HR image. Our FCGAN is an end-to-end framework, without any pre/post-processing (e.g., face alignment, extracting facial structure prior information). Furthermore, it is a robustly model, the generative image is not sensitive to facial expression, pose, illumination, occlusion (wearing glasses or hat), and so on. For the generator and discriminator networks, the skip-layer connection technique is utilized for enhancing the convergence speed in the training phase. Thus, our model has great advantages on the training time over other SR models based on CNN.

However, there are several problems that worth to further investigate in the future. We note that the input image size of recent FCGAN model is same as the generative HR image ($128\times128$). In the future research, we will design an advanced model that can directly generate HR face image (e.g., $128\times128$) from the small size one (e.g., $32\times32$). In addition, we only show the excellent performance on face image SR task in this work, and it is worth to extend our proposed framework for the task of generic image SR.

%

\ifCLASSOPTIONcompsoc
  \section*{Acknowledgments}
\else
  \section*{Acknowledgment}
\fi

This work is supported by the National Natural Science Foundation of China under Grant 73003802 and the Major International (Regional) Joint Research Project of China under Grant 73045702.



%
%
%

\bibliography{refer}

\begin{thebibliography}{10}

\bibitem{Jia2011Fast}
Zhen Jia, Hongcheng Wang, Ziyou Xiong, and A.~Finn.
\newblock Fast face hallucination with sparse representation for video
  surveillance.
\newblock In {\it Pattern Recognition}, pages 179--183, 2011.

\bibitem{zhu2016deep}
Shizhan Zhu, Sifei Liu, Chen~Change Loy, and Xiaoou Tang.
\newblock Deep cascaded bi-network for face hallucination.
\newblock In {\it European Conference on Computer Vision}, pages 614--630.
  Springer, 2016.

\bibitem{wang2014comprehensive}
Nannan Wang, Dacheng Tao, Xinbo Gao, Xuelong Li, and Jie Li.
\newblock A comprehensive survey to face hallucination.
\newblock {\it International journal of computer vision}, 106(1):9--30, 2014.

\bibitem{li-PR2014face}
Yongchao Li, Cheng Cai, Guoping Qiu, and Kin-Man Lam.
\newblock Face hallucination based on sparse local-pixel structure.
\newblock {\it Pattern Recognition}, 47(3):1261--1270, 2014.

\bibitem{autee2015review}
Ms~Prachi Autee, Mr~Samyak Mehta, Ms~Sampada Desai, Vinaya Sawant, and Anuja
  Nagare.
\newblock A review of various approaches to face hallucination.
\newblock {\it Procedia Computer Science}, 45:361--369, 2015.

\bibitem{Jiang2016Noise-TCYB}
J.~Jiang, J.~Ma, C.~Chen, X.~Jiang, and Z.~Wang.
\newblock Noise robust face image super-resolution through smooth sparse
  representation.
\newblock {\it IEEE Transactions on Cybernetics}, PP(99):1--12, 2016.

\bibitem{jin2015robust}
Yonggang Jin and Christos-Savvas Bouganis.
\newblock Robust multi-image based blind face hallucination.
\newblock {\it the IEEE Conference on Computer Vision and Pattern Recognition},
  pages 5252--5260, 2015.

\bibitem{su2016supervised}
Weng-Tai Su, Chih-Chung Hsu, Chia-Wen Lin, and Weiyao Lin.
\newblock Supervised-learning based face hallucination for enhancing face
  recognition.
\newblock In {\it 2016 IEEE International Conference on Acoustics, Speech and
  Signal Processing (ICASSP)}, pages 1751--1755. IEEE, 2016.

\bibitem{W2016DeepJFHR}
Junyu Wu, Shengyong Ding, Wei Xu, and Hongyang Chao.
\newblock Deep joint face hallucination and recognition.
\newblock {\it arXiv preprint arXiv:1611.08091}, 2016.

\bibitem{David-BEGAN2017}
David Berthelot, Tom Schumm, and Luke Metz.
\newblock {BEGAN:} boundary equilibrium generative adversarial networks.
\newblock {\it arXiv preprint arXiv:1703.10717}, 2017.

\bibitem{Christian-SRGAN-CVPR2017}
Christian Ledig, Lucas Theis, Ferenc Huszar, Jose Caballero, Andrew~P. Aitken,
  Alykhan Tejani, Johannes Totz, Zehan Wang, and Wenzhe Shi.
\newblock Photo-realistic single image super-resolution using a generative
  adversarial network.
\newblock {\it arXiv preprint arXiv:1609.04802}, 2016.

\bibitem{dong2016accelerating}
Chao Dong, Chen~Change Loy, and Xiaoou Tang.
\newblock Accelerating the super-resolution convolutional neural network.
\newblock In {\it European Conference on Computer Vision}, pages 391--407.
  Springer, 2016.

\bibitem{Dong-He-PAMI2016Image}
C.~Dong, C.~C. Loy, K.~He, and X.~Tang.
\newblock Image super-resolution using deep convolutional networks.
\newblock {\it IEEE Transactions on Pattern Analysis Machine Intelligence},
  38(2):295--307, 2016.

\bibitem{Jiang2016SRLSP}
Junjun Jiang, Chen Chen, Jiayi Ma, Zheng Wang, Zhongyuan Wang, and Ruimin Hu.
\newblock Srlsp: A face image super-resolution algorithm using smooth
  regression with local structure prior.
\newblock {\it IEEE Transactions on Multimedia}, 19(1):27--40, 2017.

\bibitem{Kim_2016_DRCN}
Jiwon Kim, Jung~Kwon Lee, and Kyoung~Mu Lee.
\newblock Deeply-recursive convolutional network for image super-resolution.
\newblock {\it The IEEE Conference on Computer Vision and Pattern Recognition
  (CVPR Oral)}, June 2016.

\bibitem{LapSRN_CVPR2017}
Wei-Sheng Lai, Jia-Bin Huang, Narendra Ahuja, and Ming-Hsuan Yang.
\newblock Deep laplacian pyramid networks for fast and accurate
  super-resolution.
\newblock In {\it IEEE Conferene on Computer Vision and Pattern Recognition},
  2017.

\bibitem{Yu-URGAN-ECCV2016}
Xin Yu and Fatih Porikli.
\newblock {\it Ultra-Resolving Face Images by Discriminative Generative
  Networks}, pages 318--333.
\newblock Springer International Publishing, Cham, 2016.

\bibitem{He2015Deep}
Kaiming He, Xiangyu Zhang, Shaoqing Ren, and Jian Sun.
\newblock Deep residual learning for image recognition.
\newblock {\it the IEEE Conference on Computer Vision and Pattern Recognition},
  pages 770--778, 2016.

\bibitem{Goodfellow2014GAN}
Ian~J. Goodfellow, Jean Pougetabadie, Mehdi Mirza, Bing Xu, David Wardefarley,
  Sherjil Ozair, Aaron Courville, and Yoshua Bengio.
\newblock Generative adversarial networks.
\newblock {\it Advances in Neural Information Processing Systems},
  3:2672--2680, 2014.

\bibitem{kingma2013auto}
Diederik~P Kingma and Max Welling.
\newblock Auto-encoding variational bayes.
\newblock {\it arXiv preprint arXiv:1312.6114}, 2013.

\bibitem{denton2015deep}
Emily~L Denton, Soumith Chintala, Rob Fergus, et~al.
\newblock Deep generative image models using a￼ laplacian pyramid of
  adversarial networks.
\newblock In {\it Advances in neural information processing systems}, pages
  1486--1494, 2015.

\bibitem{Simonyan2015Very}
Karen Simonyan and Andrew Zisserman.
\newblock Very deep convolutional networks for large-scale image recognition.
\newblock {\it International Conference on Learning Representations(ICLR)},
  2015.

\bibitem{pix2pix}
Phillip Isola, Jun{-}Yan Zhu, Tinghui Zhou, and Alexei~A. Efros.
\newblock Image-to-image translation with conditional adversarial networks.
\newblock {\it CoRR}, abs/1611.07004, 2016.

\bibitem{mirza2014conditionalGAN}
Mehdi Mirza and Simon Osindero.
\newblock Conditional generative adversarial nets.
\newblock {\it arXiv preprint arXiv:1411.1784}, 2014.

\bibitem{Arjovsky2017TowardsWGAN}
Martin Arjovsky and Léon Bottou.
\newblock Towards principled methods for training generative adversarial
  networks.
\newblock {\it arXiv preprint arXiv:1701.04862}, 2017.

\bibitem{Arjovsky2017WGAN}
Martin Arjovsky, Soumith Chintala, and Léon Bottou.
\newblock Wasserstein gan.
\newblock {\it arXiv preprint arXiv:1701.07875}, 2017.

\bibitem{ronneberger2015U-net}
Olaf Ronneberger, Philipp Fischer, and Thomas Brox.
\newblock U-net: Convolutional networks for biomedical image segmentation.
\newblock In {\it International Conference on Medical Image Computing and
  Computer-Assisted Intervention}, pages 234--241. Springer, 2015.

\bibitem{huang2016Densenet}
Gao Huang, Zhuang Liu, Kilian~Q Weinberger, and Laurens van~der Maaten.
\newblock Densely connected convolutional networks.
\newblock {\it arXiv preprint arXiv:1608.06993}, 2016.

\bibitem{orhan2017skip}
A~Emin Orhan.
\newblock Skip connections as effective symmetry-breaking.
\newblock {\it arXiv preprint arXiv:1701.09175}, 2017.

\bibitem{Densenetsemantic}
Simon J{\'{e}}gou, Michal Drozdzal, David V{\'{a}}zquez, Adriana Romero, and
  Yoshua Bengio.
\newblock The one hundred layers tiramisu: Fully convolutional densenets for
  semantic segmentation.
\newblock {\it arXiv preprint arXiv:1611.09326}, abs/1611.09326, 2016.

\bibitem{zhao2016EBGAN}
Junbo Zhao, Michael Mathieu, and Yann LeCun.
\newblock Energy-based generative adversarial network.
\newblock {\it arXiv preprint arXiv:1609.03126}, 2016.

\bibitem{liu2015Celeba}
Ziwei Liu, Ping Luo, Xiaogang Wang, and Xiaoou Tang.
\newblock Deep learning face attributes in the wild.
\newblock In {\it Proceedings of the IEEE International Conference on Computer
  Vision}, pages 3730--3738, 2015.

\end{thebibliography}

\end{document}